\documentclass[runningheads]{llncs}
\usepackage{graphicx}
\usepackage{amsmath,multirow,booktabs,tabularx}
\usepackage{textcomp}
\usepackage{array} 
\usepackage{longtable}
\usepackage{booktabs}
\usepackage{float}
\usepackage{anyfontsize}
\usepackage{color} 
\usepackage{algorithm}  
\usepackage{algorithmicx}  
\usepackage{algpseudocode}  
\usepackage{amsmath} 
\usepackage{amsfonts}
\usepackage{hyperref}

\usepackage{bm}
\usepackage{subfigure}
\usepackage{parskip}
\usepackage{amssymb}
\usepackage{CJKutf8}
\usepackage{adjustbox}
\usepackage{multirow}
\usepackage{verbatimbox}

\usepackage{latexsym}

\usepackage[T1]{fontenc}

\usepackage[utf8]{inputenc}
\usepackage{microtype}

\begin{document}
%
\title{Revisit Input Perturbation Problems for LLMs: A Unified
Robustness Evaluation Framework for Noisy Slot Filling Task\thanks{The first two authors contribute equally. Weiran Xu is the corresponding author.}}
\titlerunning{Revisit Input Perturbation Problems for LLMs}
\author{
Guanting Dong\inst{1*},
Jinxu Zhao\inst{1*},
Tingfeng Hui\inst{1},
Daichi Guo\inst{1},
Wenlong Wan\inst{1},
Boqi Feng\inst{1},
Yueyan Qiu\inst{1},
Zhuoma Gongque\inst{1},
Keqing He\inst{2},
Zechen Wang\inst{1},
Weiran Xu\inst{1*}
}
\authorrunning{G. Dong et al.}
\institute{Beijing University of Posts and Telecommunications, Beijing, China \and
Meituan Group, Beijing, China \\
\email{
\{dongguanting,zhaojinxu,huitingfeng,xuweiran\}@bupt.edu.cn}}

\maketitle              

\begin{abstract}
With the increasing capabilities of large language models (LLMs), these high-performance models have achieved state-of-the-art results on a wide range of natural language processing (NLP) tasks. However, the models' performance on commonly-used benchmark datasets often fails to accurately reflect their reliability and robustness when applied to real-world noisy data. To address these challenges, we propose a unified robustness evaluation framework based on the slot-filling task to systematically evaluate the dialogue understanding capability of LLMs in diverse input perturbation scenarios. Specifically, we construct a input perturbation evaluation dataset, Noise-LLM, which contains five types of single perturbation and four types of mixed perturbation data. Furthermore, we utilize a multi-level data augmentation method (character, word, and sentence levels) to construct a candidate data pool, and carefully design two ways of automatic task demonstration construction strategies (instance-level and entity-level) with various prompt templates. Our aim is to assess how well various robustness methods of LLMs perform in real-world noisy scenarios. The experiments have demonstrated that the current open-source LLMs generally achieve limited perturbation robustness performance. Based on these experimental observations, we make some forward-looking suggestions to fuel the research in this direction\footnote{The code is available at \href{https://github.com/dongguanting/Noise-Slot-Filling-LLM}{https://github.com/dongguanting/Noise-Slot-Filling-LLM}}.
\end{abstract}
\section{Introduction}

The slot filling (SF) task in the goal-oriented dialog system aims to identify task-related slot types in certain domains for understanding user utterances. 
Recently, Large-scale language models (LLMs) \cite{brown2020language,touvron2023llama,openai2023gpt4} have shown an impressive ability for in-context learning with only a few task-specific examples as demonstrations. Under the framework of in-context learning, LLMs have achieved promising results in a variety of NLP tasks, including machine translation (MT) \cite{reimers2019sentence}, question answering (QA) \cite{Lazaridou2022InternetaugmentedLM} and named entity extraction (NEE) \cite{brown2020language}.
However, despite its powerful abilities, the high performance of these models depend heavily on the distribution consistency between training data and test data. 
As the data distribution of dialogues in real scenarios is unknown\cite{wu2021bridging}, there are still many challenges in applying these methods to real dialogue scenarios.
\begin{figure}[t]
\centering
\resizebox{0.95\textwidth}{!}{\includegraphics{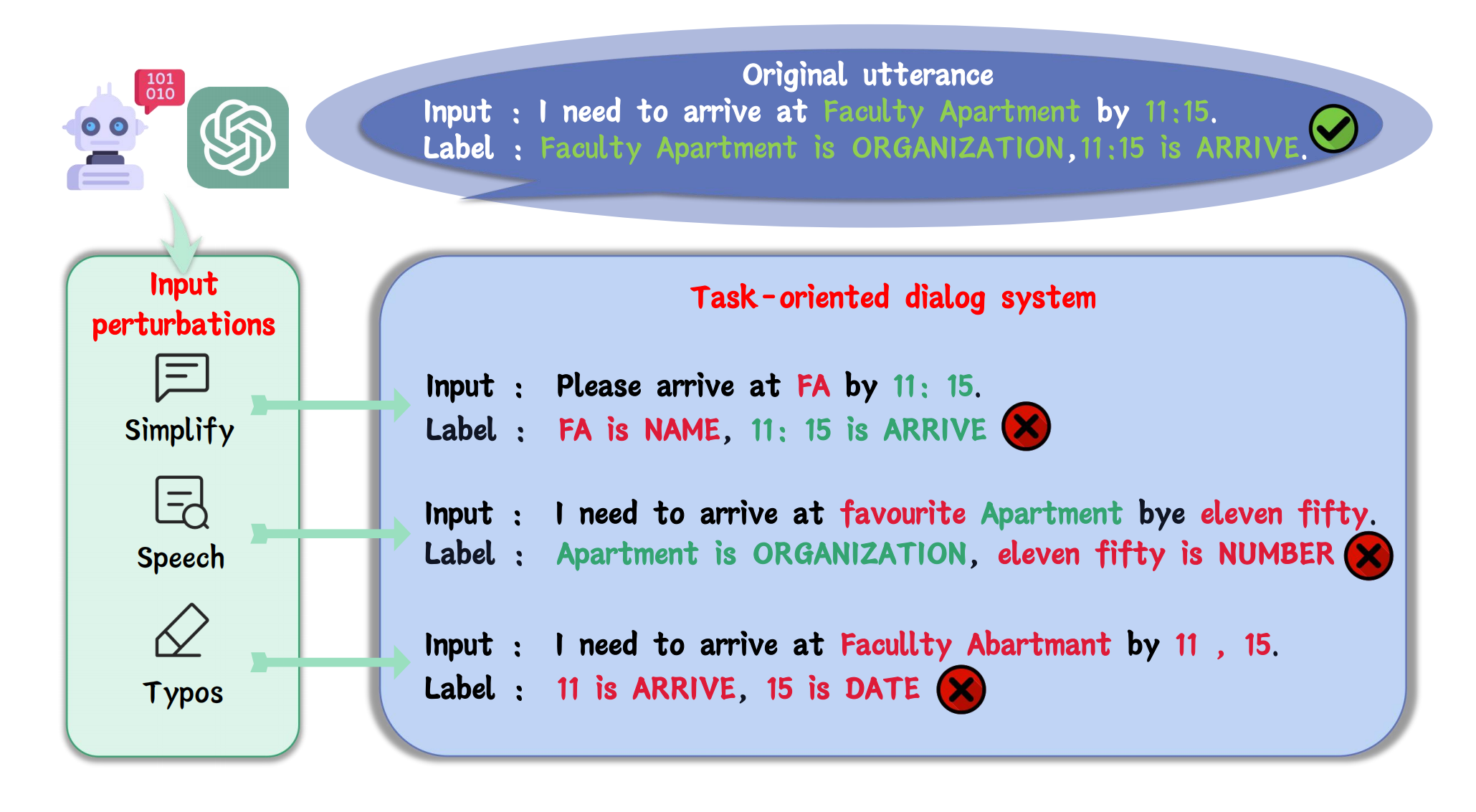}}
\vspace{-0.4cm} 
\caption{The impact of various types of input perturbations on the slot filling system in real-word scenarios}
 
\label{fig:intro}
\vspace{-0.4cm} 
\end{figure}

In real dialogue systems, due to the diverse language expression and input errors of humans, models often need to deal with a variety of input perturbations.
As shown in Figure \ref{fig:intro}, due to different expression habits, users may not interact with the dialogue system following the standard input format, but may simplify their queries to express the same intent. What's more, errors in the upstream input system may also introduce disturbances to the downstream model (e.g. typos from keyboard input, speech errors from ASR systems). 
LLMs are usually pre-trained and fine-tuned on perturbation-free datasets, resulting in poorer performance than supervised small language models with robust settings on noisy slot filling tasks. However, the accuracy of slot filling tasks directly reflects the model's understanding of user queries, which would impact the performance of the model in other downstream tasks. Therefore, exploring the robustness and generalization of LLMs under various input perturbations is crucial for the application of task-oriented dialogue systems in real scenarios.

To address the above challenges, in this paper, we aim to investigate how well various robustness methods of large language models perform in real-world noisy scenarios, and then provide empirical guidance for the research of robust LLMs. 
Based on the slot filling task, we propose a unified robustness evaluation framework to evaluate the dialogue understanding capability of large models in diverse input perturbation scenarios. Firstly, we construct a input perturbation evaluation dataset, Noise-LLM, to investigate the robustness of large language models in two different input perturbation settings. 1) \textbf{Single perturbation setting:}  Based on the DST dataset Raddle \cite{peng2020raddle} which contains various types of real-world noisy texts, we transformed them into slot filling data through manual annotation which consists of 5 types of single perturbation at the character, word, and sentence levels. 2) \textbf{Mixed perturbation setting:} We utilized the widely used Slot Filling dataset SNIPS \cite{coucke2018snips} and data augmentation tools \cite{gui2021textflint} to construct 4 types of mixed perturbation that fit real-world dialog scenarios. Furthermore, we utilize a multi-level data augmentation method (character, word, and sentence levels) to construct a candidate data pool, and carefully designed two ways of automatic task demonstration construction strategies (Instance-level and Entity-level) with various prompt templates. Our goal is to provide empirical guidance for research on robust LLMs based on our evaluation framework.
Based on this framework, we conduct extensive experiments on Noise-LLM and analyze the results from the perspectives of input perturbation types and different demonstrations, respectively. 

Our main contributions are concluded as follows:

(1) To our best knowledge,  we are the first to comprehensively investigate the effects of input perturbations at different levels on LLMs, and construct an input perturbation evaluation dataset, Noise-LLM, which includes single perturbation and mixed perturbation settings that fit real-world scenarios.

(2) We propose a unified robustness evaluation framework based on the SF task, which includes a multi-level data augmentation method, diverse prompt templates, and various automatic task demonstration construction strategies.

(3) Experiments result demonstrate that the current open-source LLMs generally have limited ability to counter input perturbations. The extensive analysis also provides empirical guidance for further research.

\section{Related Work}

\subsection{Input Perturbation Problem}
Owing to the notable performance gaps between benchmarks and real-world scenarios, the robustness of NLP systems in the face of input perturbations has garnered significant attention recently. \cite{moradi-samwald-2021-evaluating,9747192} conduct empirical evaluations of the robustness of various NLP systems against input perturbations using synthetic benchmarks that we generated. \cite{liu2023robust,dong-etal-2022-pssat,Guo2023RevisitOP} focus on the robustness of the sequence labeling framework against optical input perturbations (e.g. misspellings, OOV). \cite{gopalakrishnan2020neural} investigate the robustness of dialogue systems on ASR noise. \cite{Ruan2020} mainly focus on the ASR noise-robustness SLU models in dialogue systems. In this paper, we investigate the abilities of ChatGPT against input perturbation problems for slot filling tasks.

\begin{figure*}
    \centering
    \resizebox{1\textwidth}{!}{
    \includegraphics{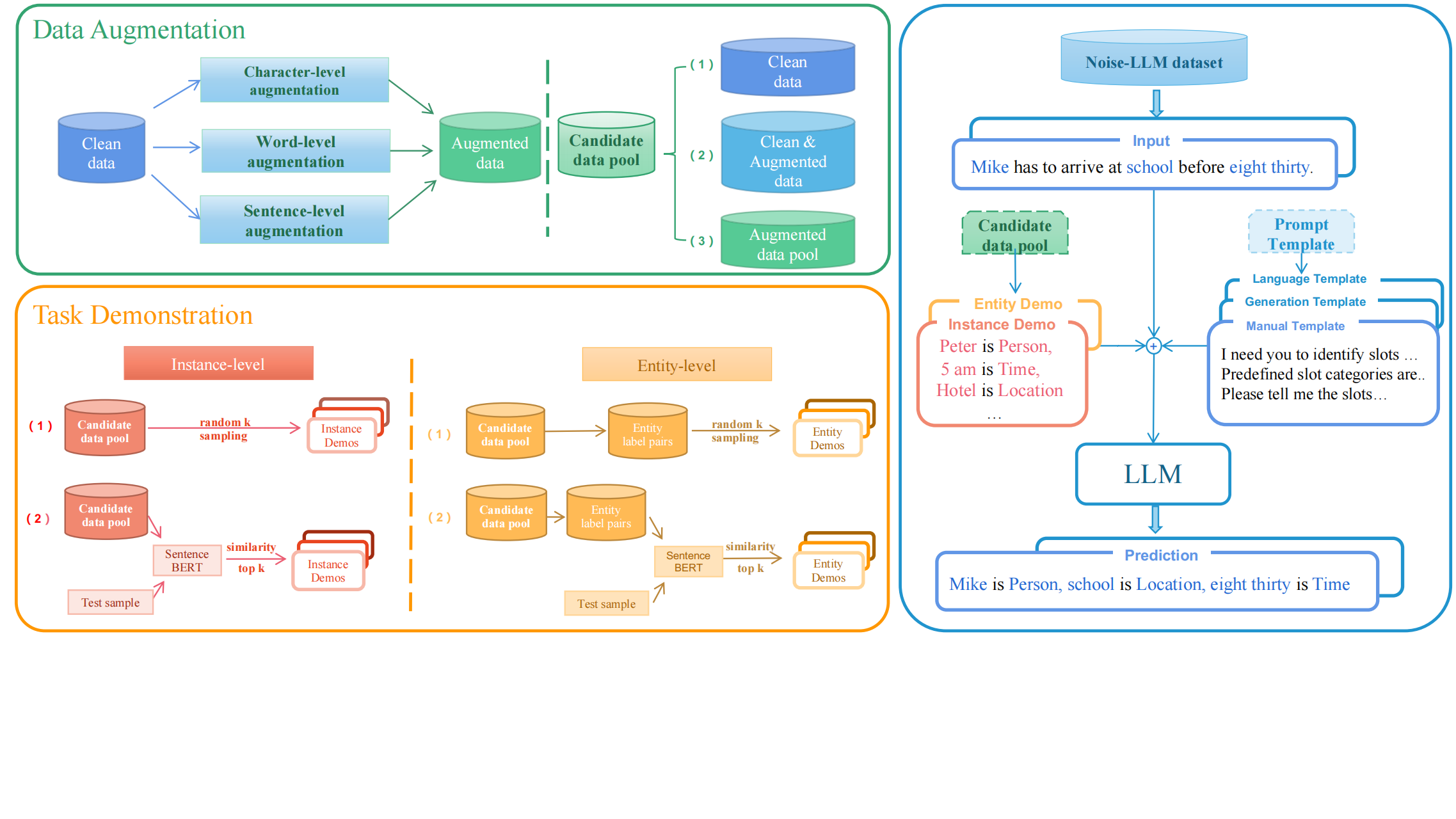}}
    \vspace{-2cm}
    \caption{The overall architecture of our unified robustness evaluation framework based on slot filling task}

    \label{fig:main}
    \vspace{-0.4cm}
\end{figure*}
\subsection{Large Language Models}
The emergence of large-scale language models (LLMs) have brought revolutionary transformation to the field of natural language processing (NLP) \cite{shen2023hugginggpt}. LLMs, such as GPT-3 \cite{brown2020language}, LLaMA \cite{touvron2023llama}, ChatGPT, and GPT-4 \cite{openai2023gpt4}, have demonstrated impressive abilities on various tasks and in generating fluent responses due to the large-scale training corpora, as well as the use of external techniques such as instruction tuning and reinforcement learning from human feedback (RLHF) \cite{ouyang2022training,hu2021lora}. LLMs based on generative framework reformulate the information extraction task \cite{lei2023instructerc,li-etal-2023-generative}, task-oriented dialog systems \cite{zeng-etal-2022-semi,qixiang-etal-2022-exploiting}, even from multi modal perspective \cite{bai2023qwen,10.1145/3447548.3467206,zhang-etal-2023-pay,zhao-etal-2022-entity}. More recently, the NLP community has been exploring various application directions for LLMs. For instance, chain-of-thought prompting \cite{wei2023chainofthought} enables LLMs to generate problem-solving processes step-by-step. Recent works \cite{yuan2023scaling,luo2023wizardmath,li2023query} integrates different LLM's decoding paths or augmented data as the supervision signal, which significantly enhancing the model's reasoning ability. Some researchers have utilized the powerful interactive capabilities of LLMs to generate commands that invoke external tools for better handling of various downstream tasks \cite{shen2023hugginggpt,lu2023instag}.  
Although DMT\cite{dong2023abilities} has begun to empirically explore the balance problem of multiple tasks using LLMs. However, there is still a blank for the input disturbance problem.
In this paper, we aim to investigate the robustness of LLMs against input perturbations. To be specific, we explore several prompt and demonstration strategies and conduct a large number of experiments to advance our research.


\section{Method}
\subsection{Problem Definition}
Given an input utterance $ X = \{ x_{1}, x_{2}, ..., x_{N}\} $, where $ N$ represents the length of $ X $, we adopt a triple $ y_{i}=\{l,r,t\} \in Y $ to represent the $ i-th $ entity that appears in $ X $, where $ Y $ represents all the entity triplets in $ X $ , and $ l, r $ denote the entity boundaries, while t denotes the entity type. We use $ D_{clean} $, $ D_{augment} $, and $ D_{test} $ to denote the conventional labeled slot filling datasets, including the clean dataset, the data-augmented dataset, and the test set, respectively, where $ D_{test} = \{X_{1}:Y_{1}, X_{2}:Y_{2}, ... X_{N}:Y_{N}\} $. We define $ P_{i} = X_{i}:Y_{i} $, and then $ D_{test} = {P_{1}, P_{2}, ... , P_{N} } $ , and we formulate the spoken language perturbation process in the real scenario as $ \widetilde{P} = \Gamma(P) $, such that $  \widetilde{X} \ne X $ but $ \widetilde{Y} $ may or may not be identical to $ Y $. The form of the perturbation transformation $\Gamma$ is flexible, which could be a specific type of perturbation (e.g., Typos) or mixed perturbation.Based on the above process, we can obtain a perturbed test set $ \widetilde{D}_{test} $.

\begin{figure*}
    \centering
    \resizebox{\textwidth}{!}{
    \includegraphics{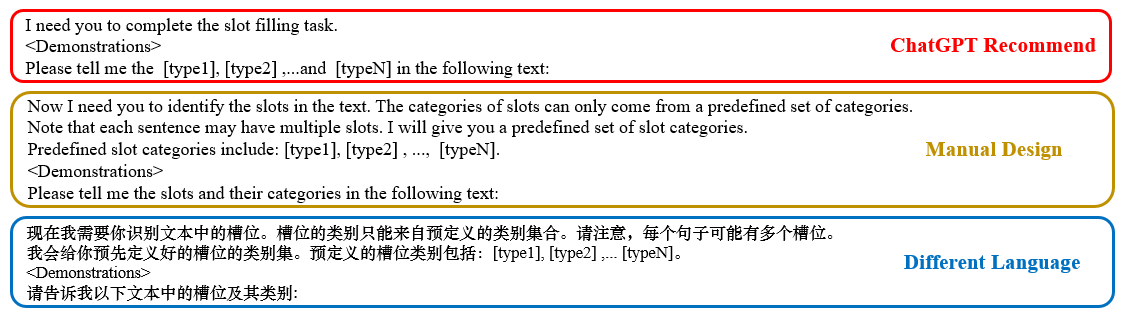}}
    \vspace{-0.6cm}
    \caption{The different prompt templates for slot filling task.}
    \label{fig:demo}
    \vspace{-0.5cm}
\end{figure*}

\subsection{Data Augmentation}
As shown in the top left corner of Figure \ref{fig:main}, we utilize the data augmentation tool NLPAug \cite{ma2019nlpaug} to transform the clean training set into an augmented dataset. The specific data augmentation methods employed in this study can be categorized into the following three types: \textbf{Character-level augmentation}: randomly add, delete and replace characters in one token with the probability $ p $.\textbf{ Word-level augmentation }: randomly delete and insert words and replace words with homophones in one sentence with probability $ p $. \textbf{ Sentence-level augmentation }: replace sentences with synonymous ones.

\subsection{Task Demonstration}
Task demonstration $ D $ is constructed by retrieving instance examples from a dataset. The role of constructing task demonstration is to select appropriate examples that can effectively demonstrate how the LLM should solve the task. We categorize the demonstration into two types: (1) Entity-oriented demonstration, and (2) Instance-oriented demonstration. 

\textbf{Entity-oriented demonstration.} Given a set of entity type labels $ L = { l_{1}, l_{2}, ... ,l_{N} } $, where $ N $ is the number of label $ L $ , we select one entity example $ e $ per label $ l $ from $ D_{train} $ and modify it into the form  [$  e_{i} $ is $ l_{i}  $]. For each label $ l $, we adopt two strategies for selecting entity $ e $: (1) Randomly selecting entity $ e $ from the clean, augmented, or mixed data pool. (2) Retrieving the $ k $ most relevant entity examples e from the clean, augmented, or mixed data pool using SentenceBERT \cite{reimers2019sentence}. Specifically, the SentenceBERT algorithm generates independent CLS embeddings for the input $ x $ and $ e $, and calculates the cosine similarity between them to rank the entity $ e $.

\textbf{Instance-oriented demonstration.} Given an example $ S $ and its label $ Y $, we modify all of its entities and their corresponding labels into the form of "$  e_{i} $ is $ l_{i}  $", and concatenate them with $ S $ to form $ S' $. We also adopt two strategies to select example $ S $: (1) Randomly selecting example $ S $ from the clean, augmented, or mixed data pool. (2) Retrieving the k most relevant examples $ S $ from the clean, augmented, or mixed data pool using SentenceBERT.

\textbf{Prompt template.} ChatGPT is a large language model that relies on prompts to guide its output generation. The quality of the output can be influenced by the style of the prompts used. Since slot filling is essentially a sequence labeling task rather than a generation task, it is not entirely consistent with the paradigm of contextual learning. Therefore, we have designed three templates specifically for slot filling.
To design these prompts, we initially sought inspiration from ChatGPT by requesting its advice. However, since the templates provided by ChatGPT were not specifically tailored to our task, we felt that this could potentially impact its performance. Hence, we also designed prompts based on the requirements of slot filling. Considering that ChatGPT is primarily pre-trained on English language data, its understanding of prompts in other languages may vary. Therefore, we also tested the English templates translated into Chinese using ChatGPT. The three prompts used in our experiments are shown in Figure \ref{fig:demo}.

\subsection{In-Context Inference}
In-context learning (ICL) is popularized as a way of learning for large language models in the original GPT-3 paper \cite{Brown2020LanguageMA}. 
When using in-context learning, we give the large language model a task description and a set of demonstration examples (input-label pairs). A test input is then appended to the end of the demonstration examples, and the LM is allowed to make predictions only based on the conditioned demonstration examples. 
In order to answer the question correctly, the model needs to understand the demonstration examples of ICL to determine the input distribution, output distribution, input-output mapping, and format. 
We concatenate the prompt, demonstration, and test input for in-context learning and use it as model input for inference

\section{Experiments}

\subsection{Datasets}
 Based on RADDLE \cite{peng2020raddle} and SNIPS \cite{coucke2018snips}, we constructed the Noise-LLM dataset, which includes two settings: single perturbation and mixed perturbation. For the single perturbation setting, RADDLE is a crowd-sourced diagnostic evaluation dataset covering a broad range of real-world noisy texts for dialog systems. We extract 30 utterances for clean data, and each type of noisy utterance (Typos, Speech, Simplify, Verbose, and Paraphrase) from RADDLE to construct the evaluation set. \textbf{Typos} are caused by non-standard abbreviations, while \textbf{Speech} arises from recognition and synthesis errors from ASR systems. \textbf{Simplification} refers to users using concise words to express their intentions, while \textbf{Verbose} refers to users using redundant words to express the same intention. \textbf{Paraphrase} is also common among users who use different words or restate the text based on their language habits. For mixed perturbation setting, based on SNIPS, we used textflint \cite{gui2021textflint} to introduce character-level perturbation \textbf{Typos}, Word-level perturbation \textbf{Speech}, sentence-level perturbation \textbf{AppendIrr}, and mixed perturbation to the test set and construct a multi-perturbation evaluation set.
\begin{table*}[htbp]
     \caption{The performance (F1 score) of the finetuned SOTA (NAT, PSSAT) and LLMs (Text-davinci-003, ChatGPT) with best entity-oriented and instance-oriented demonstrations on clean and noisy test sets. }
    \centering
    \setlength{\tabcolsep}{2.5pt}
    \renewcommand{\arraystretch}{1.3}
    \begin{adjustbox}{max width=\textwidth}
        \begin{tabular}{|l|c|c|c|ccc|c|}
            \hline  
            \multirow{2}{*}{\textbf{Methods}}& \multirow{2}{*}
            {\textbf{Clean}}&\multicolumn{1}{c}{\textbf{Character}}\vline &\multicolumn{1}{c}{\textbf{Word}}\vline &\multicolumn{3}{c}{\textbf{Sentence}}\vline &\multirow{2}{*}{\textbf{Overall}} \\
            \cline{3-7}
            
             &  & Typos &Speech&Paraphrase & Simplification &  Verbose &   \\
            \hline
            NAT($ \mathcal{L}_{aug} $) & 96.01 & 67.47 &85.23 & 87.73 & 87.32 &  85.41 &  87.21 \\
            
            NAT($ \mathcal{L}_{stabil} $) & 96.04 &67.54 &85.16 & 87.42 & 87.33 &  85.29 &  87.27 \\
            
            PSSAT & 96.42 &68.34 & 85.65 & 91.54 & 89.73 &  85.82 &  88.16 \\
            \hline
            Text-davinci-003  & 43.09 &34.26 &39.34 & 38.42 & 40.12 &  37.18 &  38.54 \\
            
            ChatGPT & 71.43 &40.65& 60.00  & 55.56 & 65.54 &  55.56 &  57.21 \\
            
            ChatGPT{\scriptsize+Instance level} & 68.21(-3.2) & 65.04(+24.3) & 70.56(+10.5) & 58.82(+2.2) & 73.02(+7.4) &  61.77(+6.2) &  68.34(+11.1) \\
            
            ChatGPT{\scriptsize+Entity level} & 74.07(+2.6) & 62.18(+21.5) & 55.39(+4.6) & 75.59(+18.9) & 70.96(+5.4) &  71.75(+16.1) &  71.55(+14.3) \\
            \hline
        \end{tabular}
    \end{adjustbox}
    \vspace{0.2cm}
    
    \label{tab:main1}
    \vspace{-0.6cm}
 
\end{table*}

\subsection{Implementation Details}
We use the default settings to invoke the OpenAI API for text-davinci-003. For ChatGPT, we manually evaluate the results using the corresponding platform and demo websites. For PSSAT \cite{dong-etal-2022-pssat}, we contact the authors and obtain the source code. We follow PSSAT's experiment settings for their upstream work and downstream work. For all the experiments of PSSAT, we conduct the training and testing stages on the A6000 GPU.




\subsection{Baselines and Large Language Models}
\textbf{NAT} \cite{namysl-etal-2020-nat} provides two perturbation-aware training methods(data augmentation \& stability training) to improve the robustness of the model against perturbations.\\

\textbf{PSSAT} \cite{dong-etal-2022-pssat} introduces two MLM-based training strategies to better learn contextual knowledge from perturbed corpus, and utilizes a consistency processing method to filter the generated data.  \\

\textbf{Text-davinci-003} \cite{brown2020language} is the most advanced GPT-3.5 model with 175B parameters. \\

\begin{table*}[htbp]
    \caption{The performance of ChatGPT with different example selection strategies on clean and noisy test sets. The two best methods for each perturbation are marked.}
    \centering
    \setlength{\tabcolsep}{2.5pt}
    \renewcommand{\arraystretch}{1.3}
    \begin{adjustbox}{max width=\textwidth}
        \begin{tabular}{|l|c|c|c|c|c|ccc|c|}
            \hline 
            \multirow{2}{*}{\textbf{Demonstration}}&\multirow{2}{*}{\textbf{Strategy}}&\multirow{2}{*}{\textbf{Setup}}& \multirow{2}{*}
            {\textbf{Clean}}&\multicolumn{1}{c}{\textbf{Character}}\vline &\multicolumn{1}{c}{\textbf{Word}}\vline &\multicolumn{3}{c}{\textbf{Sentence}}\vline &\multirow{2}{*}{\textbf{Overall}} \\
            \cline{5-9}

             & &  &  & \textbf{Typos} &\textbf{Speech} &\textbf{Paraphrase} & \textbf{Simplification} &  \textbf{Verbose} &   \\
            \hline
            \multirow{6}{*}{\centering Instance-oriented level} & 
            \multirow{3}{*}{Random} & Clean & \textbf{76.92} & 58.73 &\textbf{68.26} & 58.61 & \textbf{74.19} &  57.78 &  65.36\\
            & & Augment & 68.21 &\textbf{65.04} &\textbf{70.56} & 58.82 & \textbf{73.02} &  61.77 &  \textbf{68.34}\\
            & & Clean+Augment & 71.31 &61.01 &57.81 & 53.43 & 65.03 &  63.71 &  62.32\\
            \cline{2-10}
            & \multirow{3}{*}{Retrieve} & Clean & 56.49 & 54.84 & 60.00 & 45.53 & 53.84 &  52.55 &  54.37 \\
            &  & Augment & 61.42 &51.67 &64.66 & 51.97 & 64.00 &  61.04 &  60.25\\
            &  & Clean+Augment & 60.32 &52.44 &60.93 &  55.82 & 64.06 &  44.96 & 58.24\\
            \hline
            \multirow{6}{*}{\centering Entity-oriented level} & 
            \multirow{3}{*}{Random} & Clean & 74.07 &\textbf{62.18} &55.39 & \textbf{75.59} & 70.96 &  \textbf{71.75} &  \textbf{71.55} \\
            & & Augment & \textbf{78.39} &58.54 & 58.91 &\textbf{66.14} & 70.40 &  \textbf{64.75} &  65.61 \\
            & & Clean+Augment & 70.09 & 60.19 &61.88 &  64.12 & 65.50 &  57.37 & 63.22 \\
            \cline{2-10}
            & \multirow{3}{*}{Retrieve} & Clean & 53.66 & 52.89 &54.55 & 54.13 & 57.37 &  42.96 &  53.42 \\
            &  & Augment & 70.49 & 51.72 & 68.21 & 55.82 & 63.86 &  47.83 &  61.38\\
            &  & Clean+Augment & 72.58 & 61.16 &65.40 & 42.46 & 71.64 &  46.97 &  62.84 \\
            \hline
        \end{tabular}

    \end{adjustbox}
    \vspace{0.2cm}
    \label{tab:main2}
    \vspace{-0.5cm}
\end{table*}
\textbf{ChatGPT} is created by fine-tuning a GPT-3.5 series model via instruction tuning and reinforcement learning from human feedback (RLHF).




\subsection{Main Result}
\subsubsection{Perturbation level.}
Our experimental results are shown in Table \ref{tab:main1}. LLM performance (ChatGPT, Text-davinci-003) is far behind fintune SOTA (NAT, PSSAT) regardless of whether it is on a clean test set or with perturbations at various levels. This phenomenon may be due to the fact that large models are usually pre-trained on large-scale general training data, which makes it difficult to perform well on specific domain data in zero-shot situations.
Further comparison of ChatGPT's performance on clean data and various levels of perturbations shows that large models have the most significant drop in character-level perturbation \textbf{Typos} (71.43->40.65) and sentence-level perturbation \textbf{Verbose} (71.43->55.56), with no significant changes in \textbf{Simplification}. Based on the main experimental table and specific case analysis, we have three explanations:

1. Fine-grained perturbation like \textbf{Typos} affect the semantics of the entity itself. Although large language models can infer from the context of the sample and have some robustness in recognizing slot entities, the drastic changes in entity semantics can make it difficult to predict the correct entity label.

2. \textbf{Verbose} language tends to use redundant expressions to convey the intended meaning, often without explicitly mentioning specific entities. ChatGPT will over-predict entities due to knowledge confusion, which greatly affects its performance in the verbose domain.

3. Despite affecting the semantic information of the original sentence, \textbf{simplification} perturbation mitigates the chances of incorrect slot position prediction in ChatGPT by reducing the complexity of input content. Consequently, the model performs relatively better in the Simplification domain.

\subsubsection{Demonstration level.}
As is shown in Table \ref{tab:main2}, incorporating entity-oriented and instance-oriented demonstration strategies into ChatGPT indicates a substantial improvement in the overall results (14.34 for Entity and 11.13 for Instance), which demonstrates the effectiveness of our approach. Our evaluation primarily involved analyzing 1) the impact of different types of demonstration, 2) the selection of demonstration, and 3) the distribution of demonstration on the results.
\begin{table*}[htbp]
    \caption{ The performance of Text-davinci-003 and ChatGPT with best entity-oriented
    and instance-oriented demonstrations on mixed perturbation.}
    \centering
    \setlength{\tabcolsep}{2.5pt}
    \renewcommand{\arraystretch}{1.3}
    \begin{adjustbox}{max width=\textwidth}
        \begin{tabular}{|l|c|ccc|cccc|c|}
        \hline  
            \multirow{2}{*}{\textbf{Methods}}& \multirow{2}{*}
            {\textbf{Clean}}&\multicolumn{1}{c}{\textbf{Char}} &\multicolumn{1}{c}{\textbf{Word}} &\multicolumn{1}{c}{\textbf{Sen}} \vline&\multicolumn{1}{c}{\textbf{Char+Word}} &\multicolumn{1}{c}{\textbf{Word+Sen}}&\multicolumn{1}{c}{\textbf{Char+Sen}}&\multicolumn{1}{c}{\textbf{Char+Word+Sen}} \vline
            &\multirow{2}{*}{\textbf{Overall}} \\
            \cline{3-9}

         & &   \textbf{Typos} &  \textbf{Speech} &\textbf{AppendIrr} & \textbf{Spe+Typ} & \textbf{Spe+App} & \textbf{Ent+App} & \textbf{Spe+App+Typ} &  \\
        \hline
        Text-davince-003 & 31.24 & 27.18  & 23.41 &27.48  &  19.32 & 19.78 & 20.73 & 18.84 & 24.64 \\
        ChatGPT & 59.65 &  42.11 & 34.83  & 45.61  & 27.58 & 31.03 & 26.38 &  26.11 & 38.18 \\
        ChatGPT{\scriptsize+Instance level} & 67.18 &  48.94 & 42.25& 52.61    & 34.26 & 38.79 & 38.64 &  30.67 & 46.58 \\
        ChatGPT{\scriptsize+Entity level} & 65.71 &  47.36 & 40.37  & 53.42  & 36.55 & 37.35 & 34.21 &  29.06 & 44.27 \\
        \hline
        \end{tabular}
    \end{adjustbox}
    \vspace{0.1cm}
    \vspace{-0.4cm}
    \label{tab:mix}
\end{table*}

\subsubsection{The type of demonstration.}
Firstly, the entity-level strategy performs better than the instance-level strategy overall, and both strategies show a significant improvement on character-level perturbation (Typos). Specifically, the entity-level strategy exhibits a remarkable improvement on coarse-grained perturbation such as verbose and paraphrase. This is because entity demonstrations provide clear entity boundary information and corresponding labeling, aiding in distinguishing entities from non-entities and greatly reducing entity boundary ambiguity.
The instance-level strategy, compared to the entity-level strategy, significantly improves the model's ability to handle Simplify and Speech perturbations. This is due to the diverse context examples provided in instance demonstrations which enhance the model's understanding of domain-specific context information and the distribution of slots, assisting in better reasoning to test samples.

\subsubsection{The selection of demonstration.}
In most cases, randomly selecting samples proves to be a more effective strategy for both entity-level and instance-level methods, compared to selecting samples based on similarity. The latter may even further reduce the model's performance. One plausible explanation for this finding is that in noisy scenarios, semantic similarity is generally low (<0.4), making it challenging to ensure consistency with the test sample distribution. By contrast, randomly retrieved samples offer a diverse set of noisy samples that can positively stimulate the context learning ability of large models.

\subsubsection{The distribution of demonstration.}
Our experimental results show that, for instance-level models, selecting 10 examples exclusively from the augmented data pool leads to the highest performance on the noisy test dataset. In contrast, for entity-level models, selecting 10 examples from the clean data pool results in the best overall performance on the noisy test dataset.
We believe that instance augmented demonstrations enable LLMs to learn the slot information and the correspondence between slots and labels under input perturbations, thereby enhancing the LLMs' ability to understand semantic information in noisy input and increasing their robustness to noise. In contrast, entity demonstrations only provide the correspondence between entities and labels, and selecting examples from the augmented data pool may mislead LLMs to incorrectly identify and classify entities. Furthermore, entity clean demonstrations enable LLMs to learn the correct entity boundaries and the correspondence between entities and labels, thus achieving better performance.
However, when we select 10 examples from a mixed data pool of clean and augmented examples, ChatGPT's performance is not as good as when selecting examples exclusively from the clean or augmented data pool. In some cases, it even performs worse than ChatGPT without any examples. This also demonstrates that consistent distribution of inputs in the demonstrations contributes to performance gains during in-context learning.

%

\begin{figure}[t]
    \centering
    \subfigure[ ]{
        \includegraphics[width=.45\textwidth]{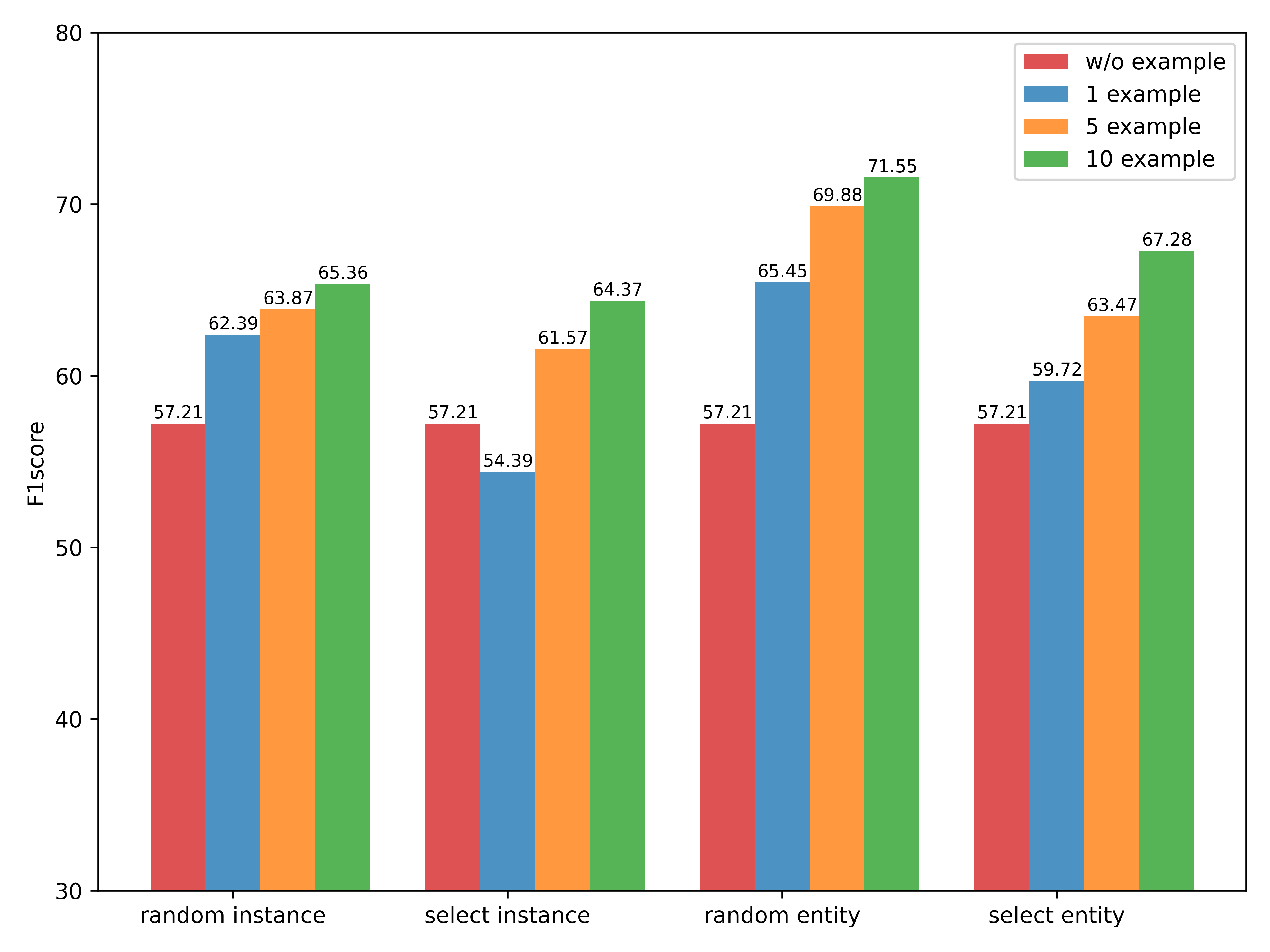}
        \label{label_for_cross_ref_1}
    }
    \subfigure[ ]{
	\includegraphics[width=.45\textwidth]{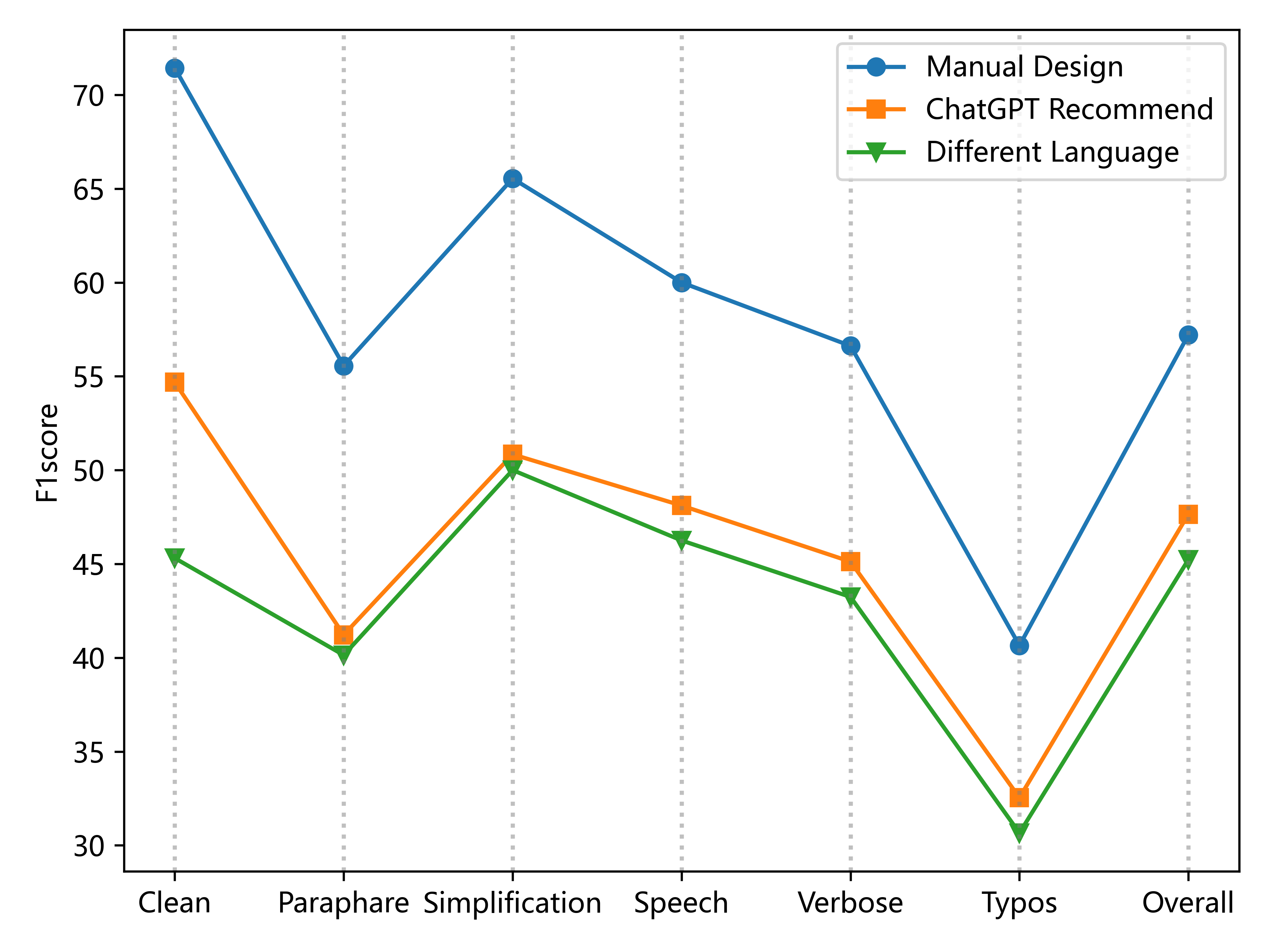}
        \label{label_for_cross_ref_2}
    }
    \vspace{-0.3cm}
    \caption{(a) The influence of the number of examples in demonstrations.
     (b) The performance of different types of prompt templates.}
     
    \label{fig:demo and prompt}
    \vspace{-0.3cm}
\end{figure}

\subsubsection{Mixed perturbation experiment.}
Table \ref{tab:mix} shows the performance of different models in facing single and mixed perturbation. Obviously, ChatGPT's performance will drop significantly when facing mixed perturbation, and is lower than all single perturbation results. Although ChatGPT performs better than Text-Davince-003, this improvement becomes very limited in the multi-perturbation scenario. We found that adding entity or instance examples to ChatGPT can effectively improve the model's perturbation robustness. Specifically, the method of adding entity examples improves the overall performance by 22\%, while adding instance examples improves it by 21\%. Even with the joint interference of three perturbations, our method of selecting examples appropriately can still maintain a 17\% improvement in performance,  which demonstrates the effectiveness and stability of our proposed approach.

\subsubsection{Number of demonstrations.}
The Figure \ref{fig:demo and prompt}(a) shows the impact of the number of examples in demonstrations on the overall performance of chatgpt in single perturbation scenarios. As the number of examples increases, chatgpt's performance improves in both clean and noisy scenarios.
Further analysis shows that demonstration selection based on similarity improves with an increase in the number of demos, as the overall semantic similarity between demos and test examples increases and their distribution consistency improves.
However, increasing the number of demonstrations through random selection does not improve performance beyond a certain point. This is because as the number of demonstrations increases, their diversity reaches a peak, and additional demonstrations do not provide new semantic information, resulting in stable performance.


\subsubsection{Type of prompts.}
The Figure \ref{fig:demo and prompt}(b) shows the impact of different types of prompts on the results. For both the clean and noisy test sets, manually designed prompts perform better than the model-recommended templates. Obviously, when we input detailed descriptions of slot filling task paradigms into LLMs, they can better understand the slot filling task and achieve better performance. When translating the manually designed templates into Chinese, the model's performance also dropped significantly, which verifies our hypothesis that ChatGPT has different levels of understanding of prompts in different languages.
ompared to manually constructed templates, ChatGPT's recommended templates show a significant difference, highlighting both the sensitivity of LLMs to templates and the challenge of using LLMs to automate template construction.

\section{Conclusion}
In this paper, we are the first to comprehensively investigate the effects of input perturbations at different levels on LLMs. We further propose a unified robustness evaluation framework for noisy slot filling to systematically evaluate the dialogue understanding capability of LLMs in diverse input perturbation scenarios. Specifically, we construct a noise evaluation dataset, Noise-LLM, which contains five types of single noise and four types of mixed noise data. 
Moreover, we utilized a multi-level data augmentation method to construct a candidate data pool, and carefully designed two ways of automatic task demonstration construction strategies with various prompt templates. Experiments result demonstrate that the current open-source LLMs generally have limited ability to counter input perturbations. Our analysis provides empirical guidance for future research.

\bibliographystyle{splncs04}
\bibliography{mybibliography}

\begin{thebibliography}{10}
\providecommand{\url}[1]{\texttt{#1}}
\providecommand{\urlprefix}{URL }
\providecommand{\doi}[1]{https://doi.org/#1}

\bibitem{bai2023qwen}
Bai, J., Bai, S., Yang, S., Wang, S., Tan, S., Wang, P., Lin, J., Zhou, C.,
  Zhou, J.: Qwen-vl: A frontier large vision-language model with versatile
  abilities. arXiv preprint arXiv:2308.12966  (2023)

\bibitem{Brown2020LanguageMA}
Brown, T.B., Mann, B., Ryder, N., Subbiah, M., Kaplan, J., Dhariwal, P.,
  Neelakantan, A., Shyam, P., Sastry, G., Askell, A., Agarwal, S.,
  Herbert-Voss, A., Krueger, G., Henighan, T.J., Child, R., Ramesh, A.,
  Ziegler, D.M., Wu, J., Winter, C., Hesse, C., Chen, M., Sigler, E., Litwin,
  M., Gray, S., Chess, B., Clark, J., Berner, C., McCandlish, S., Radford, A.,
  Sutskever, I., Amodei, D.: Language models are few-shot learners. ArXiv
  \textbf{abs/2005.14165} (2020)

\bibitem{brown2020language}
Brown, T.B., Mann, B., Ryder, N., Subbiah, M., Kaplan, J., Dhariwal, P.,
  Neelakantan, A., Shyam, P., Sastry, G., Askell, A., Agarwal, S.,
  Herbert-Voss, A., Krueger, G., Henighan, T., Child, R., Ramesh, A., Ziegler,
  D.M., Wu, J., Winter, C., Hesse, C., Chen, M., Sigler, E., Litwin, M., Gray,
  S., Chess, B., Clark, J., Berner, C., McCandlish, S., Radford, A., Sutskever,
  I., Amodei, D.: Language models are few-shot learners (2020)

\bibitem{coucke2018snips}
Coucke, A., Saade, A., Ball, A., Bluche, T., Caulier, A., Leroy, D., Doumouro,
  C., Gisselbrecht, T., Caltagirone, F., Lavril, T., Primet, M., Dureau, J.:
  Snips voice platform: an embedded spoken language understanding system for
  private-by-design voice interfaces (2018)

\bibitem{dong-etal-2022-pssat}
Dong, G., Guo, D., Wang, L., Li, X., Wang, Z., Zeng, C., He, K., Zhao, J., Lei,
  H., Cui, X., Huang, Y., Feng, J., Xu, W.: {PSSAT}: A perturbed semantic
  structure awareness transferring method for perturbation-robust slot filling.
  In: Proceedings of the 29th International Conference on Computational
  Linguistics. pp. 5327--5334. International Committee on Computational
  Linguistics, Gyeongju, Republic of Korea (Oct 2022),
  \url{https://aclanthology.org/2022.coling-1.473}

\bibitem{dong2023abilities}
Dong, G., Yuan, H., Lu, K., Li, C., Xue, M., Liu, D., Wang, W., Yuan, Z., Zhou,
  C., Zhou, J.: How abilities in large language models are affected by
  supervised fine-tuning data composition (2023)

\bibitem{gopalakrishnan2020neural}
Gopalakrishnan, K., Hedayatnia, B., Wang, L., Liu, Y., Hakkani-Tur, D.: Are
  neural open-domain dialog systems robust to speech recognition errors in the
  dialog history? an empirical study (2020)

\bibitem{gui2021textflint}
Gui, T., Wang, X., Zhang, Q., Liu, Q., Zou, Y., Zhou, X., Zheng, R., Zhang, C.,
  Wu, Q., Ye, J., Pang, Z., Zhang, Y., Li, Z., Ma, R., Fei, Z., Cai, R., Zhao,
  J., Hu, X., Yan, Z., Tan, Y., Hu, Y., Bian, Q., Liu, Z., Zhu, B., Qin, S.,
  Xing, X., Fu, J., Zhang, Y., Peng, M., Zheng, X., Zhou, Y., Wei, Z., Qiu, X.,
  Huang, X.: Textflint: Unified multilingual robustness evaluation toolkit for
  natural language processing (2021)

\bibitem{Guo2023RevisitOP}
Guo, D., Dong, G., Fu, D., Wu, Y., Zeng, C., Hui, T., Wang, L., Li, X., Wang,
  Z., He, K., Cui, X., Xu, W.: Revisit out-of-vocabulary problem for slot
  filling: A unified contrastive frameword with multi-level data augmentations.
  ArXiv  \textbf{abs/2302.13584} (2023)

\bibitem{hu2021lora}
Hu, E.J., Shen, Y., Wallis, P., Allen-Zhu, Z., Li, Y., Wang, S., Wang, L.,
  Chen, W.: Lora: Low-rank adaptation of large language models (2021)

\bibitem{Lazaridou2022InternetaugmentedLM}
Lazaridou, A., Gribovskaya, E., Stokowiec, W., Grigorev, N.: Internet-augmented
  language models through few-shot prompting for open-domain question
  answering. ArXiv  \textbf{abs/2203.05115} (2022)

\bibitem{lei2023instructerc}
Lei, S., Dong, G., Wang, X., Wang, K., Wang, S.: Instructerc: Reforming emotion
  recognition in conversation with a retrieval multi-task llms framework. arXiv
  preprint arXiv:2309.11911  (2023)

\bibitem{li2023query}
Li, C., Yuan, Z., Dong, G., Lu, K., Wu, J., Tan, C., Wang, X., Zhou, C.: Query
  and response augmentation cannot help out-of-domain math reasoning
  generalization (2023)

\bibitem{9747192}
Li, X., Lei, H., Wang, L., Dong, G., Zhao, J., Liu, J., Xu, W., Zhang, C.: A
  robust contrastive alignment method for multi-domain text classification. In:
  ICASSP 2022 - 2022 IEEE International Conference on Acoustics, Speech and
  Signal Processing (ICASSP). pp. 7827--7831 (2022).
  \doi{10.1109/ICASSP43922.2022.9747192}

\bibitem{li-etal-2023-generative}
Li, X., Wang, L., Dong, G., He, K., Zhao, J., Lei, H., Liu, J., Xu, W.:
  Generative zero-shot prompt learning for cross-domain slot filling with
  inverse prompting. In: Findings of the Association for Computational
  Linguistics: ACL 2023. pp. 825--834. Association for Computational
  Linguistics, Toronto, Canada (Jul 2023),
  \url{https://aclanthology.org/2023.findings-acl.52}

\bibitem{10.1145/3447548.3467206}
Lin, J., Men, R., Yang, A., Zhou, C., Zhang, Y., Wang, P., Zhou, J., Tang, J.,
  Yang, H.: M6: Multi-modality-to-multi-modality multitask mega-transformer for
  unified pretraining. In: Proceedings of the 27th ACM SIGKDD Conference on
  Knowledge Discovery Data Mining. p. 3251–3261. KDD '21, Association for
  Computing Machinery, New York, NY, USA (2021). \doi{10.1145/3447548.3467206},
  \url{https://doi.org/10.1145/3447548.3467206}

\bibitem{liu2023robust}
Liu, J., Wang, L., Dong, G., Song, X., Wang, Z., Wang, Z., Lei, S., Zhao, J.,
  He, K., Xiao, B., Xu, W.: Towards robust and generalizable training: An
  empirical study of noisy slot filling for input perturbations (2023)

\bibitem{lu2023instag}
Lu, K., Yuan, H., Yuan, Z., Lin, R., Lin, J., Tan, C., Zhou, C., Zhou, J.:
  Instag: Instruction tagging for analyzing supervised fine-tuning of large
  language models (2023)

\bibitem{luo2023wizardmath}
Luo, H., Sun, Q., Xu, C., Zhao, P., Lou, J., Tao, C., Geng, X., Lin, Q., Chen,
  S., Zhang, D.: Wizardmath: Empowering mathematical reasoning for large
  language models via reinforced evol-instruct (2023)

\bibitem{ma2019nlpaug}
Ma, E.: Nlp augmentation. https://github.com/makcedward/nlpaug (2019)

\bibitem{moradi-samwald-2021-evaluating}
Moradi, M., Samwald, M.: Evaluating the robustness of neural language models to
  input perturbations. In: Proceedings of the 2021 Conference on Empirical
  Methods in Natural Language Processing. pp. 1558--1570. Association for
  Computational Linguistics, Online and Punta Cana, Dominican Republic (Nov
  2021). \doi{10.18653/v1/2021.emnlp-main.117},
  \url{https://aclanthology.org/2021.emnlp-main.117}

\bibitem{namysl-etal-2020-nat}
Namysl, M., Behnke, S., K{\"o}hler, J.: {NAT}: Noise-aware training for robust
  neural sequence labeling. In: Proceedings of the 58th Annual Meeting of the
  Association for Computational Linguistics. pp. 1501--1517. Association for
  Computational Linguistics, Online (Jul 2020).
  \doi{10.18653/v1/2020.acl-main.138},
  \url{https://aclanthology.org/2020.acl-main.138}

\bibitem{openai2023gpt4}
OpenAI: Gpt-4 technical report (2023)

\bibitem{ouyang2022training}
Ouyang, L., Wu, J., Jiang, X., Almeida, D., Wainwright, C.L., Mishkin, P.,
  Zhang, C., Agarwal, S., Slama, K., Ray, A., Schulman, J., Hilton, J., Kelton,
  F., Miller, L., Simens, M., Askell, A., Welinder, P., Christiano, P., Leike,
  J., Lowe, R.: Training language models to follow instructions with human
  feedback (2022)

\bibitem{peng2020raddle}
Peng, B., Li, C., Zhang, Z., Zhu, C., Li, J., Gao, J.: Raddle: An evaluation
  benchmark and analysis platform for robust task-oriented dialog systems.
  arXiv preprint arXiv:2012.14666  (2020)

\bibitem{qixiang-etal-2022-exploiting}
Qixiang, G., Dong, G., Mou, Y., Wang, L., Zeng, C., Guo, D., Sun, M., Xu, W.:
  Exploiting domain-slot related keywords description for few-shot cross-domain
  dialogue state tracking. In: Proceedings of the 2022 Conference on Empirical
  Methods in Natural Language Processing. pp. 2460--2465. Association for
  Computational Linguistics, Abu Dhabi, United Arab Emirates (Dec 2022),
  \url{https://aclanthology.org/2022.emnlp-main.157}

\bibitem{reimers2019sentence}
Reimers, N., Gurevych, I.: Sentence-bert: Sentence embeddings using siamese
  bert-networks. arXiv preprint arXiv:1908.10084  (2019)

\bibitem{Ruan2020}
Ruan, W., Nechaev, Y., Chen, L., Su, C., Kiss, I.: Towards an asr error robust
  spoken language understanding system. In: Interspeech 2020 (2020),
  \url{https://www.amazon.science/publications/towards-an-asr-error-robust-spoken-language-understanding-system}

\bibitem{shen2023hugginggpt}
Shen, Y., Song, K., Tan, X., Li, D., Lu, W., Zhuang, Y.: Hugginggpt: Solving ai
  tasks with chatgpt and its friends in huggingface (2023)

\bibitem{touvron2023llama}
Touvron, H., Lavril, T., Izacard, G., Martinet, X., Lachaux, M.A., Lacroix, T.,
  Rozière, B., Goyal, N., Hambro, E., Azhar, F., Rodriguez, A., Joulin, A.,
  Grave, E., Lample, G.: Llama: Open and efficient foundation language models
  (2023)

\bibitem{wei2023chainofthought}
Wei, J., Wang, X., Schuurmans, D., Bosma, M., Ichter, B., Xia, F., Chi, E., Le,
  Q., Zhou, D.: Chain-of-thought prompting elicits reasoning in large language
  models (2023)

\bibitem{wu2021bridging}
Wu, D., Chen, Y., Ding, L., Tao, D.: Bridging the gap between clean data
  training and real-world inference for spoken language understanding. arXiv
  preprint arXiv:2104.06393  (2021)

\bibitem{yuan2023scaling}
Yuan, Z., Yuan, H., Li, C., Dong, G., Tan, C., Zhou, C.: Scaling relationship
  on learning mathematical reasoning with large language models. arXiv preprint
  arXiv:2308.01825  (2023)

\bibitem{zeng-etal-2022-semi}
Zeng, W., He, K., Wang, Z., Fu, D., Dong, G., Geng, R., Wang, P., Wang, J.,
  Sun, C., Wu, W., Xu, W.: Semi-supervised knowledge-grounded pre-training for
  task-oriented dialog systems. In: Proceedings of the Towards Semi-Supervised
  and Reinforced Task-Oriented Dialog Systems (SereTOD). pp. 39--47.
  Association for Computational Linguistics, Abu Dhabi, Beijing (Hybrid) (Dec
  2022), \url{https://aclanthology.org/2022.seretod-1.6}

\bibitem{zhang-etal-2023-pay}
Zhang, Y., Wang, S., Li, P., Dong, G., Wang, S., Xian, Y., Li, Z., Zhang, H.:
  Pay attention to implicit attribute values: A multi-modal generative
  framework for {AVE} task. In: Findings of the Association for Computational
  Linguistics: ACL 2023. pp. 13139--13151. Association for Computational
  Linguistics, Toronto, Canada (Jul 2023),
  \url{https://aclanthology.org/2023.findings-acl.831}

\bibitem{zhao-etal-2022-entity}
Zhao, G., Dong, G., Shi, Y., Yan, H., Xu, W., Li, S.: Entity-level interaction
  via heterogeneous graph for multimodal named entity recognition. In: Findings
  of the Association for Computational Linguistics: EMNLP 2022. pp. 6345--6350.
  Association for Computational Linguistics, Abu Dhabi, United Arab Emirates
  (Dec 2022), \url{https://aclanthology.org/2022.findings-emnlp.473}

\end{thebibliography}
%




\end{document}